\documentclass[conference]{IEEEtran}
\ifCLASSINFOpdf
\else
\fi
\usepackage[cmex10]{amsmath}
\usepackage{amssymb}
\usepackage{framed}
\usepackage{graphicx}
\usepackage{caption}
\usepackage{subcaption}
\usepackage{subfig}

\usepackage[font=footnotesize]{caption}
\divide\pdfpxdimen by 300

\usepackage[]{algorithm2e}
\usepackage{verbatim}
\usepackage{color}
\definecolor{purple}{rgb}{0.4,.1,.9}
\definecolor{red}{rgb}{1,0,0}

\begin{document}
%
\title{A Nonparametric Framework for Quantifying Generative Inference on Neuromorphic Systems}


\author{\IEEEauthorblockN{Ojash Neopane$^{* \dagger \ddagger 1}$, Srinjoy Das$^{* \dagger \ddagger 1}$,
Ery Arias-Castro$^{*}$, Ken Kreutz-Delgado$^{\dagger \ddagger}$}

\medskip
Email: \{oneopane, s2das, eariasca, kreutz, \}@ucsd.edu
\IEEEauthorblockA{$^\dagger$ ECE, $^*$Dept. of Mathematics \& $^\ddagger$Calit2/QI Pattern Recognition Laboratory, UC San Diego, La Jolla, CA 92093\\ 
\emph{$^1$ Ojash Neopane and Srinjoy Das have contributed equally to this work.}
}
}
%


\maketitle

\begin{abstract}
Restricted Boltzmann Machines and Deep Belief Networks
have been successfully used in probabilistic generative model applications
such as image occlusion removal, pattern completion and
motion synthesis. Generative inference in such algorithms can be performed
very efficiently on hardware using a Markov Chain Monte Carlo
procedure called Gibbs sampling, where stochastic samples are drawn from noisy integrate and fire neurons implemented
on neuromorphic substrates. Currently, no satisfactory
metrics exist for evaluating the generative performance of
such algorithms implemented on high-dimensional data for neuromorphic platforms. This paper
demonstrates the application of nonparametric goodness-of-fit testing to both quantify the
generative performance as well as provide decision-directed
criteria for choosing the parameters of the neuromorphic Gibbs sampler
and optimizing usage of hardware resources used during sampling.
\end{abstract}


%
\IEEEpeerreviewmaketitle

\section{Introduction}
Restricted Boltzmann Machines (RBMs) and Deep Belief Networks (DBNs) (Fig. 1) are
stochastic graphical models that are commonly used in both discriminative and
generative applications such as image classification,  sequence completion, and speech recognition.  An RBM is a symmetrically connected,
bipartite Markov Random Field (MRF) which is composed of
neuron-like units-- these partitions are called the visible layer $v$ and
the hidden layer $h$.  In an RBM, both inference and learning use a
Markov Chain Monte Carlo (MCMC) procedure called Gibbs sampling \cite {haykin08neural}.  To implement Gibbs sampling using binary RBMs  each neuron is sampled based on a sigmoidal activation function which is a function of the total input from other connected neurons.  A DBN can be thought of as
multiple RBMs `stacked' onto each other, forming an $N$-partitioned Markov
Random Field with one visible layer $v$, and multiple hidden layers
$h_i$ with $i\in1,\dots,N-1$.

RBMs and DBNs are commonly implemented on high-performance platforms such as CPUs and GPUs 
deployed on cloud computing infrastructures.  However recent
progress in the field of neuromorphic VLSI using both digital and analog circuit elements (\cite {indiveri2011neuromorphic}, \cite{merolla2014million})  has demonstrated the potential of performing inference on RBMs/DBNs implemented on such substrates for ultra low-power, realtime applications. In such implementations, which consist of massively parallel arrays of spiking neurons, classification accuracy is an experimentally quantifiable metric for discriminative RBMs/DBNs as one can  experimentally  compute classification precision and accuracy using standard labeled datasets such as MNIST. On the other hand, implementing metrics for generative applications on such hardware remains a difficult task because of the high-dimensional nature of real-world data.\footnote{Essentially one is attempting to verify the correctness of an RBM-based probability distribution model for describing a high-dimensional data vector.}
In this paper, we demonstrate the applicability of using a highly scaleable nonparametric framework to quantify the performance of RBMs and DBNs for generative tasks on real-world high-dimensional data sets based on goodness-of-fit testing.

 \begin{figure}[!tb]
 \centering
 \begin{subfigure}{0.45\linewidth}
   \centering
   \includegraphics[width=3.5cm,scale=3]{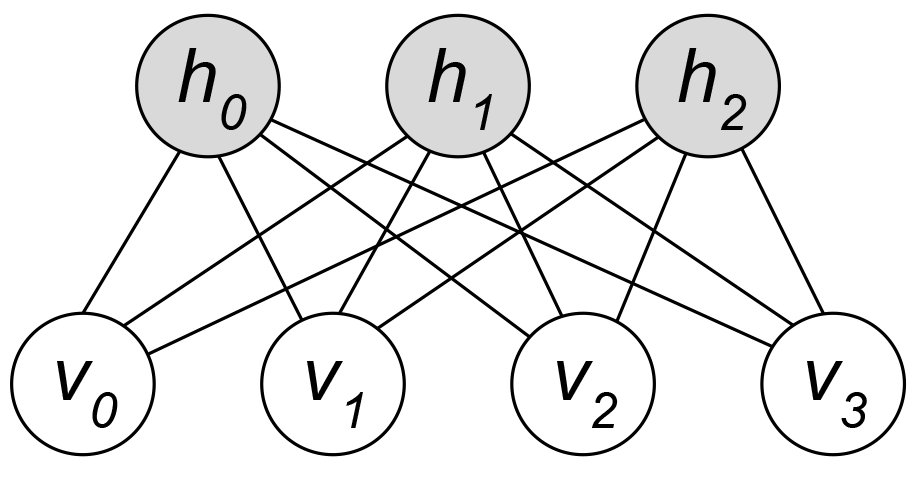}
   	\caption{}
 \end{subfigure}%
 \hfill
 \begin{subfigure}{0.45\linewidth}
   \centering
   \includegraphics[width=1.65cm,scale=3]{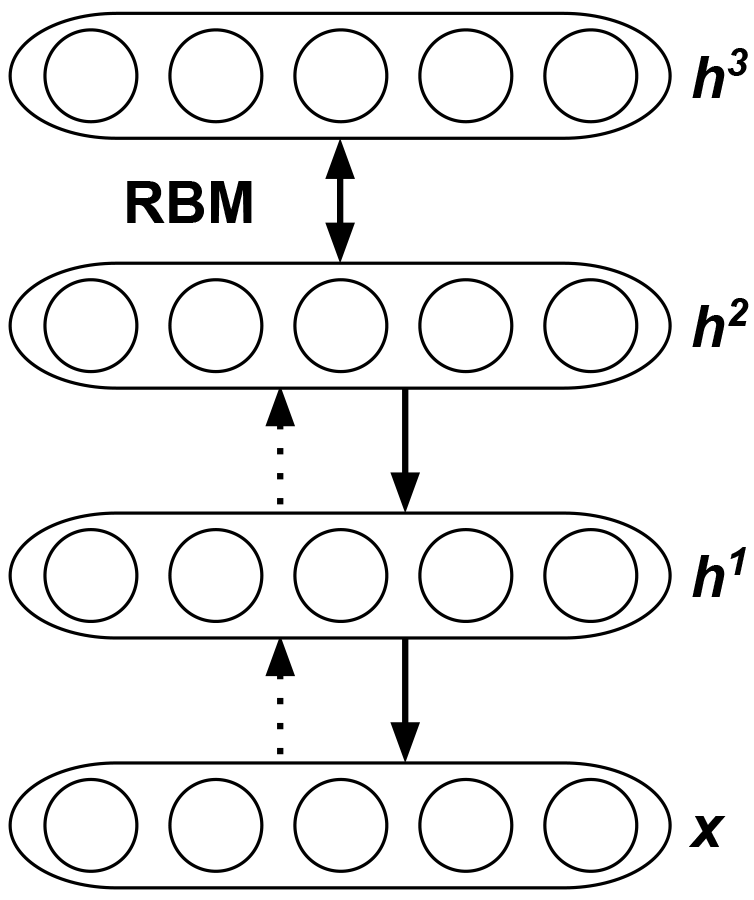}
   	\caption{}
 \end{subfigure}
 \medskip
 \medskip
 \label{RBMDNBN}
 \caption{a) Restricted Boltzmann Machine with 4 visible and 3 hidden units. b) Deep Belief Network with 3 hidden layers} 
 \end{figure}
\section{Generative Inference and Performance Metrics}

An RBM represents the probabilistic model of a data set using the
Boltzmann distribution shown below: 
\begin{equation}\begin{gathered}
p(v, h) = \frac {\exp(-E(v, h))}{\sum _{v', h'} \exp(-E(v',h'))} \\
\text{where }\ E(v,h) = -v^\top Wh - b_{\rm v}^\top v - b_{\rm h}^\top h
\label{eq:BD}
\end{gathered}\end{equation}
$E$ is often called the energy function and depends on the
state of both the visible and hidden units of the RBM. $W$ is a matrix of weights between $v$ and $h$, and $b_{\rm v}$, $b_{\rm h}$ are the biases of the visible units $v$  and hidden units $h$ respectively. For binary RBMs, the Gibbs sampling procedure guarantees that a stationary Boltzmann distribution is achieved if the state of each RBM neuron is sampled according to the sigmoidal probability rule given below \cite{haykin08neural}:
\begin{equation}
P(x_{i} = 1 | x_{j}, j \neq i) = \frac {1} {1 + \exp(-\sum_{j} w_{ij}x_{j} + b_{i}) }
\label{eq:sigmoid}
\end{equation}
Here $w_{ij}$, the $i,j$-th element of the matrix $W$, is the weight between neurons $x_{i}$ and $x_{j}$, and $b_{i}$  denotes the bias of neuron $x_{i}$.

We would like to compare the Boltzmann distribution generated by the (neuromorphic hardware) Gibbs sampler implemented on the neuromorphic substrate with that of the distribution generated by the benchmark ideal Gibbs sampler (in our case, implemented in software). 
This calls for a measure of dissimilarity between two distributions together with a way to estimate this dissimilarity based on samples from these distributions.
The Kullback-Leibler (KL) divergence \cite{cover2012elements} is a common choice.
However it is difficult to compute because doing so for RBMs requires the computation of the normalizing constant in the denominator of \eqref{eq:BD}, known as the partition function, whose explicit calculation is prohibitive for high dimensional real world data.  

Annealed Importance sampling (AIS) \cite{salakhutdinov2008quantitative} is another
algorithm that  has been used to quantify performance of RBMs/DBNs during generative tasks.   
Unlike KL-divergence, AIS does not suffer from computational complexity limitations. However, AIS only takes into consideration the parameters of the RBM in \eqref{eq:BD} trained with software and not the parameters of the hardware Gibbs sampler per se.  

A practical framework is to perform offline training of the RBM/DBN and then perform inference using Gibbs sampling by utilizing the dynamics of the underlying hardware sampler implemented on the substrate.  In this paper, we propose using  goodness-of-fit testing to compare the probability distributions generated by the hardware and benchmark software Gibbs samplers.  To illustrate the methodology, we picked the Crossmatch test \cite{rosenbaum2005exact} among the class of graph-based tests nicely reviewed in \cite{bhattacharya2015power}.
\section{The Crossmatch test}
The Crossmatch test \cite{rosenbaum2005exact} is an example of nonparametric goodness-of-fit test in arbitrary dimensions based on a graph construction.  
\def\d{d}  
Given two set of samples, $X_1, \dots, X_n$ and $Y_1, \dots, Y_n$, respectively from two probability distributions $P$ and $Q$ on some space ${\cal Z}$ equipped with a dissimilarity measure $\d : {\cal Z} \times {\cal Z} \mapsto \mathbb{R}_+$.
(The test below generalizes to the case of samples of different sizes.)
Let $Z_i = X_i$ if $i \le m$ and $= Y_{i-n}$ if $i > n$.  
The Crossmatch procedure starts by computing a optimal non-bipartite matching of the $Z$'s, meaning, it finds a permutation $\hat\sigma$ of $\{1, \dots, 2n\}$ that minimizes
\def\Def{\stackrel{\rm def}{=}}
\[
{\rm Match}(\sigma) \Def \sum_{i=1}^{2n} \d(Z_i, Z_{\sigma(i)}).
\]
We note that other matchings (such as greedy matching) can be used \cite{arias2015consistency}.  
The procedure then computes the crossmatch statistic, denoted $A$ which is defined as the number of matched pairs including an $X$ and a $Y$. 
The test rejects for {\rm small} values of $A$.  The null distribution of $A$ is known in closed form \cite{rosenbaum2005exact}:
\[\begin{gathered}
f(a) \Def P(A = a) = \frac{2^{a} n!}{ {2n \choose n} [(n-a)/2)!]^2 a!} \\
\text{for } a = 0, \dots, n.
\end{gathered}\]
Having computed $A = a_{\rm obs}$, the resulting p-value is therefore equal to $F(a_{\rm obs})$, where 
\[
F(a) \Def P(A \le a) = \sum_{a'=0}^a f(a').
\]

In our case we are dealing with distributions on $\{0,1\}^r$, where $r$ is the dimension of the visible layer.  The most common  measure of dissimilarity is the Hamming distance, meaning
\[\begin{gathered}
d(z, z') = \sum_{j=1}^r \mathbb{I}\{z_j \ne z'_j\} \\
z = (z_1, \dots, z_r) \text{ and } z' = (z'_1, \dots, z'_r) \text{ in } \{0,1\}^r.
\end{gathered}\]
This is the dissimilarity measure that we use.

Our process for comparing the output from two Gibbs samplers is as follows.  We generate $n$ samples from each Gibbs sampler and compute the p-value from the Crossmatch test based on the Hamming distance.  We repeat the process many times, resulting in that many p-values.  If the Gibbs samplers were to generate the same distribution, then the p-values would be approximately uniform in $[0,1]$.
We can then look at statistics of these p-values.  We chose to look at the mean p-value, which we use below as a similarity measure between two Gibbs sampler distributions.
{\begin{itemize}
\item A p-value near 0 indicates that the distributions are far enough for the Crossmatch test to notice with confidence that samples generated by the two distributions do not come from the same distribution.
\item A p-value near 0.5 indicates that the distributions are close enough that the Crossmatch test would not notice a (statistically) significant difference between how the two samples are distributed.
\end{itemize}

\section{Digital Sampler Parameter Selection}
TrueNorth
\  is a neuromorphic processor composed of digital integrate-and-fire (I\&F)
neurons with the capability to implement both stochastic and deterministic
leak and threshold values \cite{merolla2014million}. The following algorithm
for realizing the sigmoidal sampling rule \eqref{eq:sigmoid} using these dynamical properties  to perform MCMC sampling in RBMs  was proposed in \cite{das2015gibbs} and is as follows:



\begin{algorithm}

\textbf{Parameters:} $T_w, V_t, T_M, {\rm leak}$\\
\textbf{Input:} $v_{i} = scale*(\sum_{j} w_{ij}x_{j} + b_{i}), V = V_{initial} = v_{i}$ \\
 \Repeat{$T_w$ steps}{
$V = V + {\rm leak}*(B (0.5))$, where $B$ is Bernoulli \\
$V_{t\_rand} = V_t + \lfloor U(0,2^{T_M}-1) \rfloor$, where $U$ is Uniform \\
${\rm spiked}(V \ge V_{t\_rand}) = 1 $\\
 }
\end{algorithm}
The membrane potential of an I\&F neuron which implements an RBM neuron is denoted by $V$. sampling in this algorithm is dependent on four parameters: $T_w$, the number of
time steps used for sampling; $V_t$, the deterministic  threshold; $T_M$,
the number of bits allocated for the stochastic threshold variation; and
the value of the leak. In order to obtain a useful dynamic range, a multiplicative $scale$ factor is applied to the weights
and biases obtained from the offline RBM training \cite{das2015gibbs}. After integration, the sampled value of an RBM neuron is set to 1 if the corresponding I\&F neuron spikes in any of the allowed number of sampling intervals $T_w$. 

\begin{table}[!htb]
  \begin{center}
    \begin{tabular}{| l || c |r|}
    \hline
    Index &  (Tw, Vt, TM, leak) & scale\\
    \hline \hline
    G1 & (1, -130, 8, 0) & 50 \\
    G2 & (1, -80, 8, 102) & 50\\
    G3 & (2, 0, 8, 100) & 50 \\
    G4 & (8, 79, 9, 49) & 50 \\
    G5 & (16, 50, 9, 15) & 30\\
    G6 & (16, 100, 10, 30) & 50\\
    G7 & (16, 633, 8, 90) & 100 \\
            \hline
    \end{tabular}
  \end{center}
  \caption{Digital neuron parameters}
  \label{table1}
\end{table}
\medskip
In this paper we select 7 different sets of parameters for the digital sampler as shown in Table~\ref{table1}.
The procedure for testing the 7 samplers was as follows.  First,  we trained an RBM in software with 784 visible units and 500 hidden units on the MNIST dataset consisting of $28 \times 28$ grayscale images of 5000 handwritten digits (the training data) to generate the weight matrix as well as the bias values.  Following this, we initialized the visible units of the RBM with an initial set of values $x\in\{0,1\}^{784}$
each of which corresponded to one of the 1000 MNIST digits in the set of test data.
 We generated samples using the benchmark ideal Gibbs sampler in software which we refer to as the ideal data (see Fig. 2a).  Following this, we then use the dynamics of the 7 digital samplers to perform MCMC sampling and thereby create seven sets of samples.  Figure \ref{images} illustrates the MNIST digits obtained by  sampling with the ideal Gibbs sampler as well as with  2 hardware samplers from Table~\ref{table1} (G2 and G5).

\begin{figure}[htb]
\centering
\begin{subfigure}{0.25\linewidth}
  \centering
  \includegraphics[width=1.9cm,scale=3]{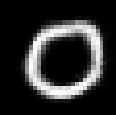}
  \caption{}
  \label{fig:ideal}
\end{subfigure}%
\hfill
\begin{subfigure}{0.25\linewidth}
  \centering
  \includegraphics[width=1.9cm,scale=3]{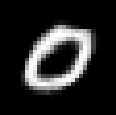}
  \caption{}
  \label{fig:g2}
\end{subfigure}
\hfill
\begin{subfigure}{0.25\linewidth}
  \centering
  \includegraphics[width=1.9cm,scale=3]{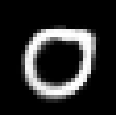}
  \caption{}
  \label{fig:g6}
\end{subfigure}
\hfill
\smallskip
\medskip
\medskip
\caption{Images generated by (a) the benchmark ideal Gibbs sampler (b) the G2 hardware sampler (c)  the G5 hardware sampler.  It is difficult to visually distinguish which hardware sampler has the best performance.}
\label{images}
\end{figure}
\smallskip
\smallskip

For comparing each digital sampler against the ideal benchmark, 5,000 trials of Crossmatch were run, resulting in 5,000 p-values (for each digital sampler).
The distribution of the p-values from the simulations are shown in Figure \ref{figure_pval}.
From this figure and the description of each digital sampler in Table~\ref{table1}, it can be seen that the sampler latency $T_w$ and $scale$ are both critical to the generative performance of the RBM. Figure~\ref{figure_testdata} shows a histogram of the distribution of p-values for digital samplers G2 and G5 (see Table~\ref{table1}). To choose the appropriate sampler out of the samplers from Table \ref{table1}, we took the ratio between the mean p-value for each sampler, and its estimated energy consumption.  The ratio of p-value to energy, the Energy Performance Efficiency (EPEff), is shown in Figure \ref{pPower}.  The figure clearly indicates that out of all samplers from Table \ref{table1} sampler G2 has the highest EPEff and is thus the appropriate sampler to choose.
\begin{figure}[!htb]
\centering
\includegraphics[width=7cm]{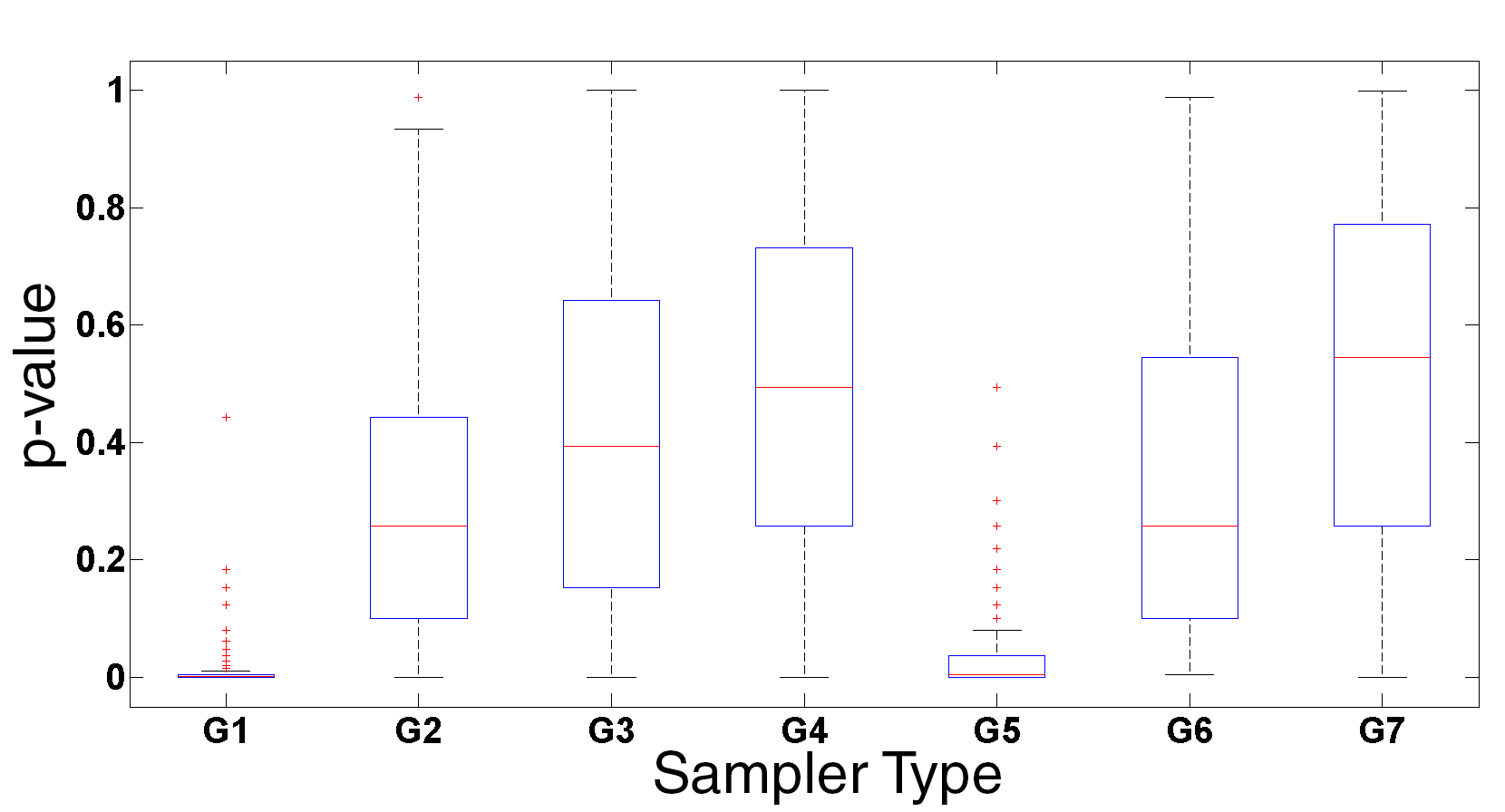}
\caption{The distribution of Crossmatch p-values for each of digital samplers in Table~\ref{table1}.  A higher mean p-value indicates a stronger statistical closeness to the ideal software Gibbs sampler}
\label{figure_pval}
\end{figure}
\begin{figure}[!htb]
\centering
\includegraphics[width=8cm]{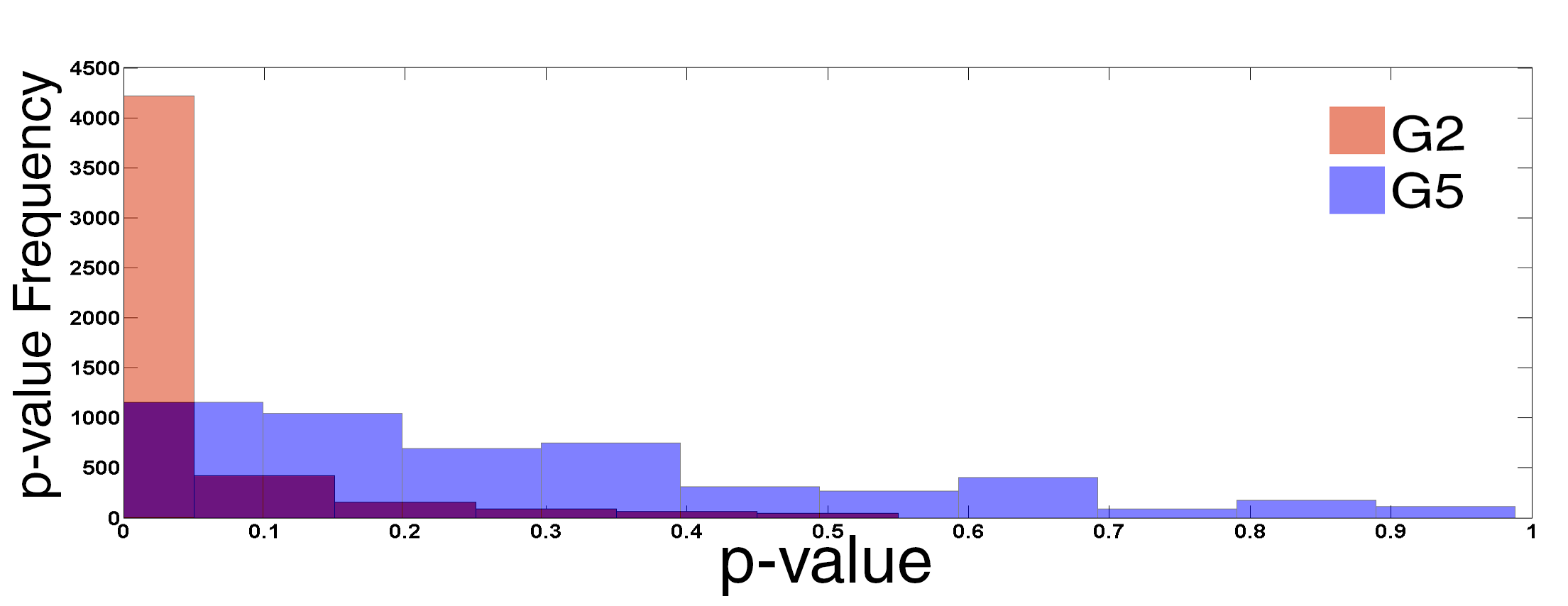}
\caption{Histogram of p-values generated from the Crossmatch test, comparing the Ideal Sampler with the G2 and G5 samplers.  The p-value distribution of the G2 sampler more closely approximates a Uniform[0,1] distribution then the G5 sampler, and is thus has the better performance of the two samplers.}
\label{figure_testdata}
\end{figure}
\begin{figure}[!htb]
\centering
\includegraphics[width=8cm]{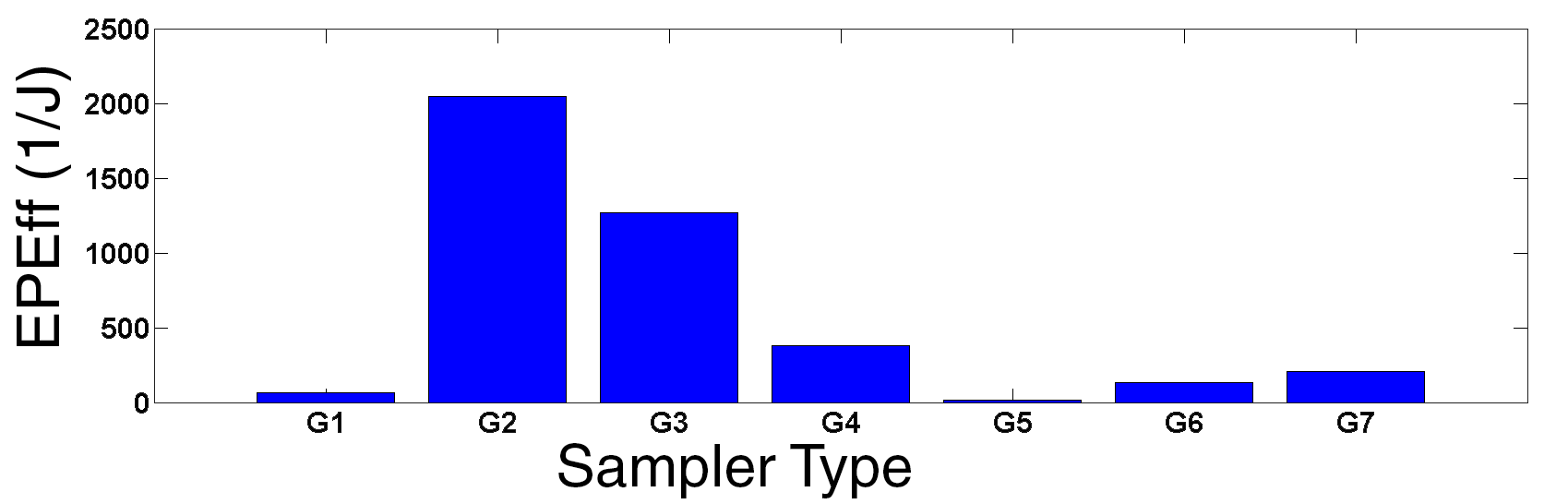}
\caption{A plot showing the estimated Energy Performance Efficiency of each of the 7 samplers}
\label{pPower}
\end{figure}

\section{Network Resource Optimization}
In order to implement the sampling algorithm described in Section IV, with arbitrary $leak$ values,  each RBM neuron, which has accumulated its input from other connected neurons, utilizes two TrueNorth neurons. Specifically, one I\&F
neuron with a stochastic threshold receiving inputs (data neuron) and a second neuron to produce the stochastic leak of 0 or 1 which is multiplied by
the weight factor on the connected axon as per the ${leak}$ value used in the algorithm (leak neuron).
Figure \ref{TN} shows this implementation where the green neurons are the
data neurons and the red neurons are the leak neurons. Since a single TrueNorth core has only 256 neurons,
a direct implementation of this mapping scheme will result in only 50 percent of the neurons on a crossbar utilized
for sampling which can result in a large number of cores for RBM/DBN implementations.
\begin{figure}[!htb]
\centering
\begin{subfigure}{0.25\linewidth}
  \centering
  \includegraphics[width=1.6cm,scale=3]{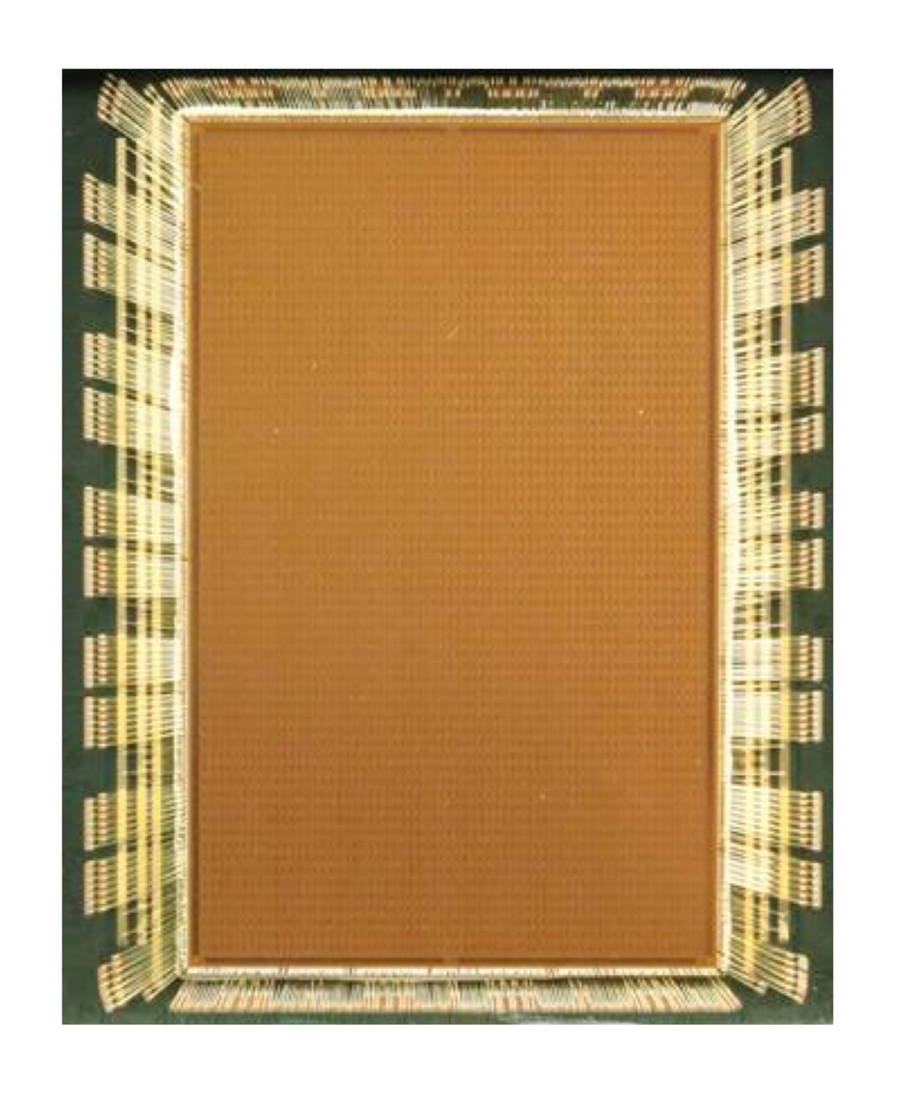}
  \caption{}
\end{subfigure}%
\hfill
\begin{subfigure}{0.25\linewidth}
  \centering
  \includegraphics[width=2.65cm,scale=3]{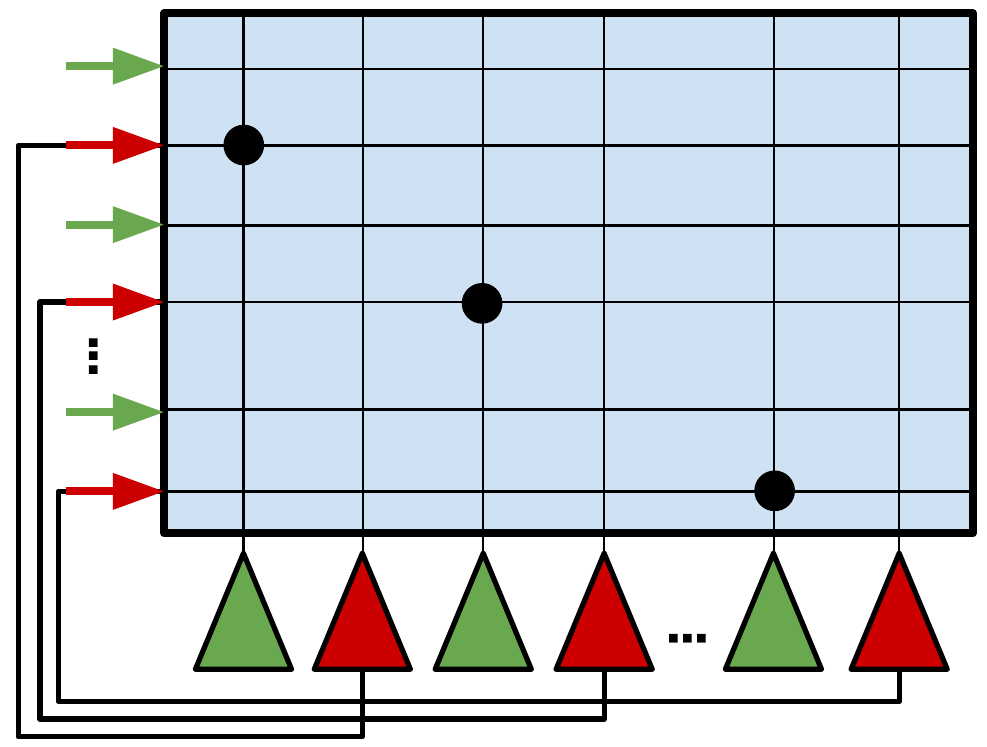}
  \caption{}
\end{subfigure}
\hfill
\begin{subfigure}{0.25\linewidth}
  \centering
  \includegraphics[width=2.65cm,scale=3]{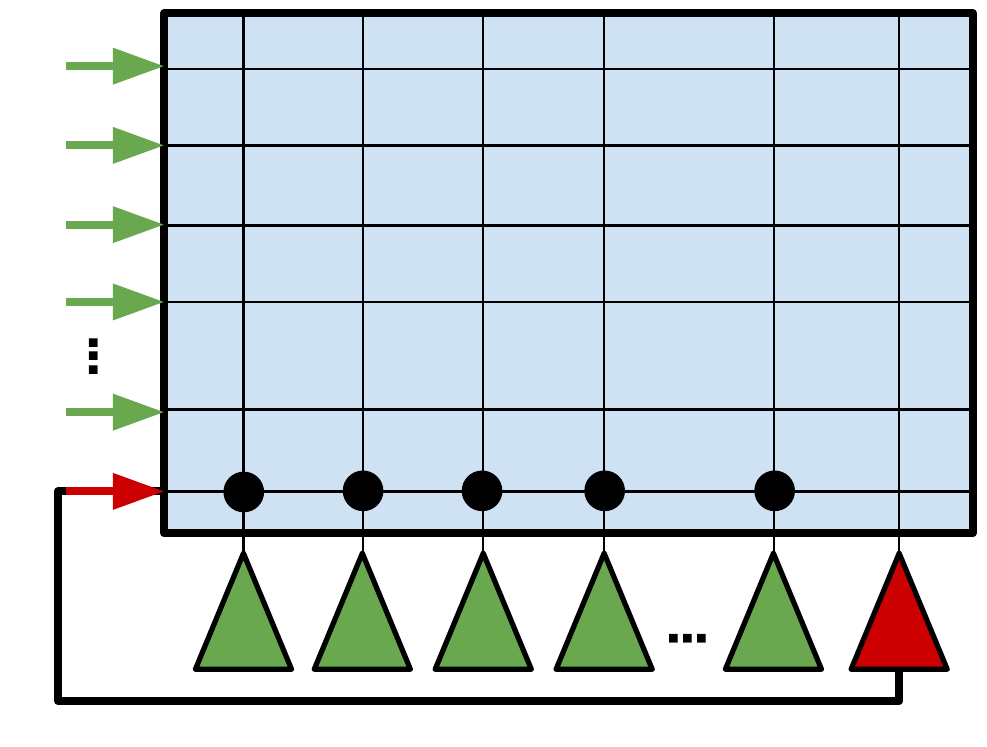}
  \caption{}
\end{subfigure}
\hfill

\medskip
\medskip
\caption{(a) The TrueNorth neuromorphic processor. (b) A diagram of a TrueNorth RBM sampling network with a leak density of one. Green neurons represent data neurons while red neurons represent leak neurons. (c) A diagram of a TrueNorth RBM sampling network with a leak density of 255.}
\label{TN}
\end{figure}

One way to improve core utilization efficiency is to limit the number of leak
neurons by relaying a single leak value to multiple data
neurons. Such a modification would however reduce the quality of the generated
samples since the original algorithm proposed in \cite{das2015gibbs} requires the ${leak}$ used for data neurons to
be identically and independently distributed (i.i.d.). In such a scenario, Crossmatch can be used to
study the loss of generative performance and determine an optimal density of data neurons
to leak neurons which is henceforth referred to as the leak density ($l_d$).

We chose sampler G2 to test the loss of performance and the optimal value of $l_d$. Using this sampler,
and 7 different values of leak density $l_d = 2, 5, 10, 50, 100, 200, 255$, we studied the variation of generative performance during sampling from the MNIST RBM implemented on TrueNorth.  
The results of the experiment are shown in Figure \ref{figure_ld} below.  


\begin{figure}[ht]
\centering
\includegraphics[width=7cm]{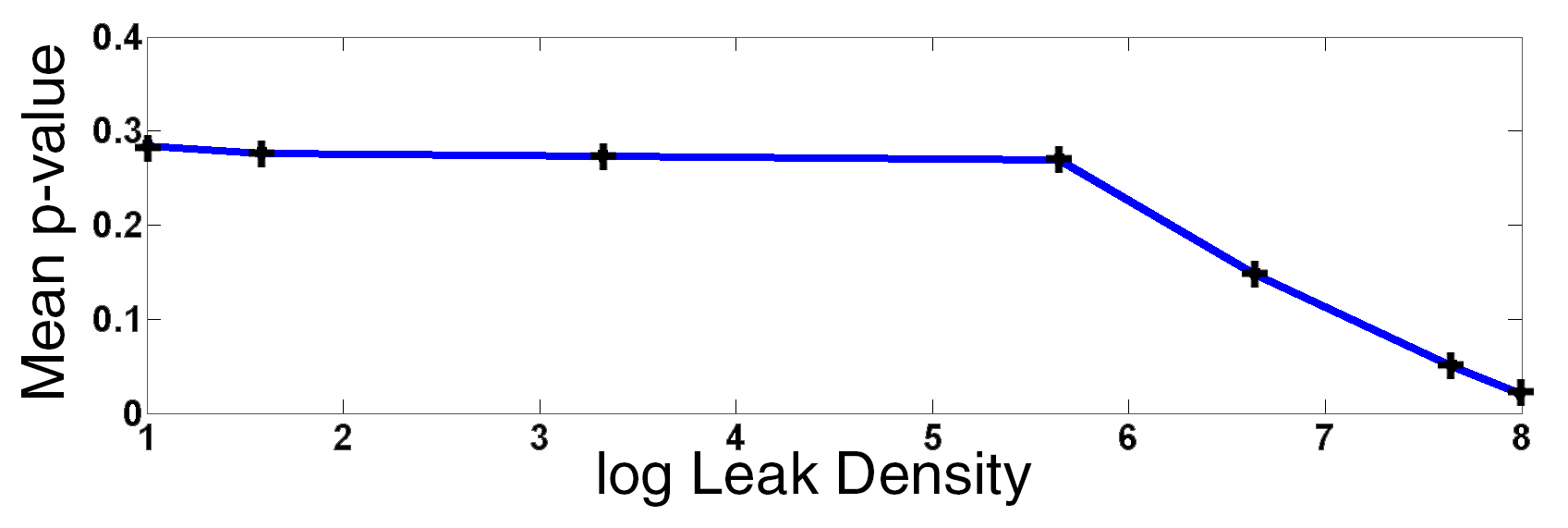}
\caption{Mean p-values for the G2 sampler with different leak density values. A higher mean p-value indicates a stronger statistical closeness to the G2 sampler with $l_d = 1$.}
\label{figure_ld}
\end{figure}
\smallskip
\smallskip
\section{Energy-performance tradeoff for Network Resource Optimization}
\newcommand{\pp}{Energy Performance Efficiency }
We studied the
variation of the EPEff (defined in Section IV) with respect to $l_d$ for the generative RBM model of MNIST implemented on TrueNorth. Here the energy estimated value consumed by TrueNorth during sampling.
The results are shown in Figure \ref{figure_power}. 
For low $l_d$ values, lower EPEff values were produced on account of the large number of cores required for sampling.  For large $l_d$ values, low EPEff values were observed on account of the significant loss in accuracy as reflected by the Crossmatch p-values.  It is clear from Figure 8 that $l_d = 10$ maximizes the EPEff, and is thus the optimal choice for implementing sampling from the MNIST RBM on TrueNorth. 
\begin{figure}[ht]
\centering
\includegraphics[width=7cm]{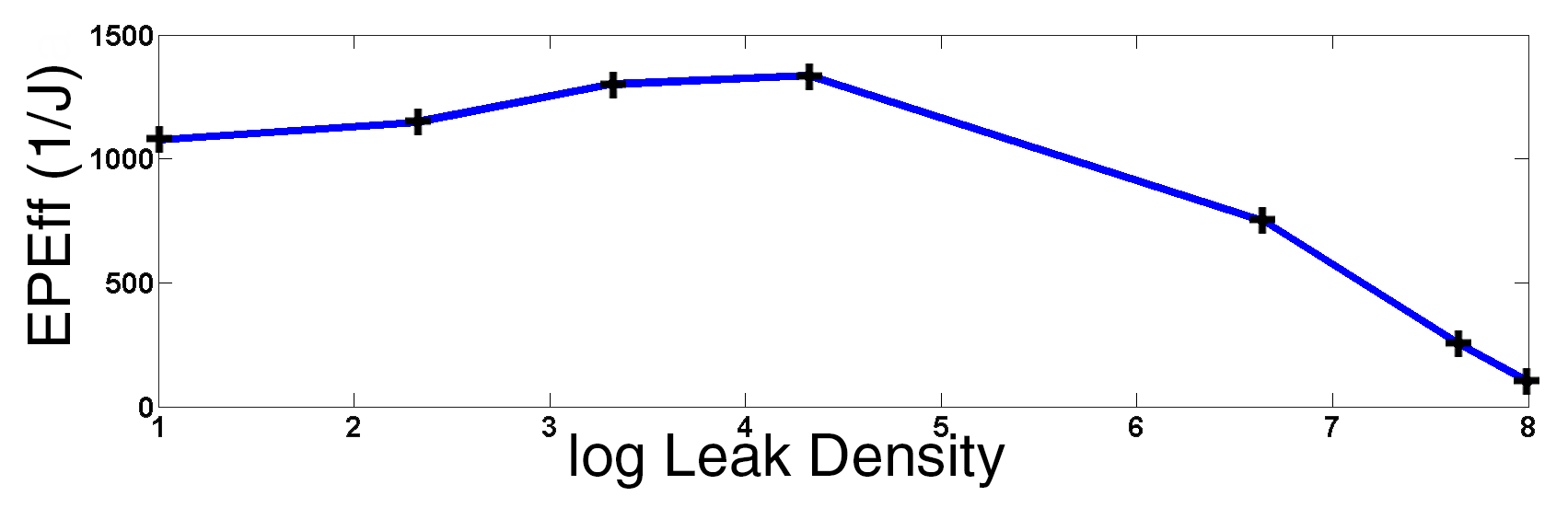}
\caption{Estimated Energy Performance Efficiency for various leak densities.  A leak density of 20 maximizes the performance of the sampler with respect to energy usage.}
\label{figure_power}
\end{figure}
\section{Applicability to design in analog neuromorphic systems}
Leaky integrate-and-fire neurons which can be implemented in analog neuromorphic systems also have 
the capability to perform Gibbs sampling from the generative model of an RBM using their own neuron dynamics \cite{neftci2013event}.  The behavior 
of such neurons are governed by the following equation:
\begin{align}
 C \frac{d}{dt}u_i = -g_Lu_i + I_i(t) + \sigma\xi(t)
\quad u_i(t)\in (V_{reset},\theta)
\label{leaky_if}
\end{align}
where C is the membrane capacitance, $V_{reset}$ is the reset potential, $u_i$ is the neuron's membrane potential,
$g_L$ is the leak conductance, $I_i(t)$ is the neuron's synaptic current,
$\theta$ is the neuron's threshold,  and $\sigma\xi(t)$ is a noise term in which $\sigma$ is
the noise variance.  Similar to the discussion in Section IV, Crossmatch can be used to determine
optimal design parameters for the Gibbs sampler on the analog neuromorphic substrate. 
Additionally in such a system each neuron has its own i.i.d.
noise term $\sigma\xi(t)$ as shown in  (\ref{leaky_if}). Such continuous-time noise injection 
utilizes significant system bandwidth hence it is highly desirable to reduce this by connecting the same noise source to 
multiple neurons.  Therefore, by applying the decision-directed strategy presented in Sections V and VI, the same noise source could be re-utilized among multiple neurons, improving system throughput, latency, power and hardware usage.

\section{Conclusions}
We have demonstrated how a nonparametric goodness-of-fit test like the Crossmatch can be used to evaluate the performance of a generative RBM model implemented on neuromorphic VLSI substrates.  Such a test, and the p-values it provides when applied to many realized samples, can provide a useful tool for quantifying the accuracy of a Gibbs sampler, whether digital or analog. In particular, for the problem at hand, such a tool can facilitate the choice of hardware parameters and optimization of network resources.  

\section{Acknowledgments}
The authors would like to thank the team members of the Brain-Inspired Computing
group  at IBM Almaden for supporting this project. The authors would  also like to
thank the Calit2/QI Pattern Recognition Laboratory at UCSD for providing funding 
support for attending the conference as well as Bruno Pedroni at UCSD.



%
\bibliography{srinjoy}

\begin{thebibliography}{10}
\providecommand{\url}[1]{#1}
\csname url@samestyle\endcsname
\providecommand{\newblock}{\relax}
\providecommand{\bibinfo}[2]{#2}
\providecommand{\BIBentrySTDinterwordspacing}{\spaceskip=0pt\relax}
\providecommand{\BIBentryALTinterwordstretchfactor}{4}
\providecommand{\BIBentryALTinterwordspacing}{\spaceskip=\fontdimen2\font plus
\BIBentryALTinterwordstretchfactor\fontdimen3\font minus
  \fontdimen4\font\relax}
\providecommand{\BIBforeignlanguage}[2]{{%
\expandafter\ifx\csname l@#1\endcsname\relax
\typeout{** WARNING: IEEEtran.bst: No hyphenation pattern has been}%
\typeout{** loaded for the language `#1'. Using the pattern for}%
\typeout{** the default language instead.}%
\else
\language=\csname l@#1\endcsname
\fi
#2}}
\providecommand{\BIBdecl}{\relax}
\BIBdecl

\bibitem{haykin08neural}
S.~Haykin, \emph{{Neural Networks and Learning Machines (3rd Edition)}}.\hskip
  1em plus 0.5em minus 0.4em\relax Prentice Hall, 2008.

\bibitem{indiveri2011neuromorphic}
G.~Indiveri, B.~Linares-Barranco, T.~J. Hamilton, A.~Van~Schaik,
  R.~Etienne-Cummings, T.~Delbruck, S.-C. Liu, P.~Dudek, P.~H{\"a}fliger,
  S.~Renaud \emph{et~al.}, ``Neuromorphic silicon neuron circuits,''
  \emph{Frontiers in neuroscience}, vol.~5, 2011.

\bibitem{merolla2014million}
P.~A. Merolla, J.~V. Arthur, R.~Alvarez-Icaza, A.~S. Cassidy, J.~Sawada,
  F.~Akopyan, B.~L. Jackson, N.~Imam, C.~Guo, Y.~Nakamura \emph{et~al.}, ``A
  million spiking-neuron integrated circuit with a scalable communication
  network and interface,'' \emph{Science}, vol. 345, no. 6197, pp. 668--673,
  2014.

\bibitem{cover2012elements}
T.~M. Cover and J.~A. Thomas, \emph{Elements of information theory}.\hskip 1em
  plus 0.5em minus 0.4em\relax John Wiley \& Sons, 2012.

\bibitem{salakhutdinov2008quantitative}
R.~Salakhutdinov and I.~Murray, ``On the quantitative analysis of deep belief
  networks,'' in \emph{Proceedings of the 25th international conference on
  Machine learning}.\hskip 1em plus 0.5em minus 0.4em\relax ACM, 2008, pp.
  872--879.

\bibitem{rosenbaum2005exact}
P.~R. Rosenbaum, ``An exact distribution-free test comparing two multivariate
  distributions based on adjacency,'' \emph{Journal of the Royal Statistical
  Society: Series B (Statistical Methodology)}, vol.~67, no.~4, pp. 515--530,
  2005.

\bibitem{bhattacharya2015power}
B.~B. Bhattacharya, ``Power of graph-based two-sample tests,'' \emph{arXiv
  preprint arXiv:1508.07530}, 2015.

\bibitem{arias2015consistency}
E.~Arias-Castro and B.~Pelletier, ``On the consistency of the crossmatch
  test,'' \emph{arXiv preprint arXiv:1509.05790}, 2015.

\bibitem{das2015gibbs}
S.~Das, B.~Pedroni, P.~Merolla, J.~Arthur, A.~Cassidy, D.~Modha,
  G.~Cauwenberghs, and K.~Kreutz-Delgado, ``{``Gibbs Sampling with Low-Power
  Spiking Digital Neurons"},'' in \emph{IEEE Int. Symp. Circuits and Systems},
  2015.

\bibitem{neftci2013event}
E.~Neftci, S.~Das, B.~Pedroni, K.~Kreutz-Delgado, and G.~Cauwenberghs,
  ``Event-driven contrastive divergence for spiking neuromorphic systems,''
  \emph{Frontiers in Neuroscience}, vol.~7, p. 272, 2013.

\end{thebibliography}
\bibliographystyle{IEEEtran}

\end{document}